\documentclass[runningheads]{llncs}
\usepackage{graphicx}
\usepackage{amsfonts,amssymb}
\usepackage{multirow}
\usepackage{color}
\usepackage{subfigure}
\usepackage[table]{xcolor}
\usepackage{hyperref}
\hypersetup{
    hidelinks
}
\usepackage{caption, subcaption}
\usepackage{marvosym}
\begin{document}

\definecolor{pos_green}{RGB}{0,176,80}
\definecolor{T1_blue}{RGB}{223, 237, 244}
\definecolor{T1_green}{RGB}{181, 231, 172} 

\title{Evaluating Attribute Comprehension in Large Vision-Language Models}
%
%

\author{Haiwen Zhang\inst{1} \and
Zixi Yang\inst{1} \and
Yuanzhi Liu\inst{1} \and
Xinran Wang\inst{1}\and
Zheqi He\inst{2} \and
Kongming Liang*\inst{1}  \and
Zhanyu Ma\inst{1}}
\authorrunning{F. Author et al.}
%
\institute{Beijing University of Posts and Telecommunications, Beijing, China \\
\email{liangkongming@bupt.edu.cn}\\ \and
Beijing Academy of Artificial Intelligence, Beijing, China
}

\maketitle              
\begin{abstract}
Currently, large vision-language models have gained promising progress on many downstream tasks. However, they still suffer many challenges in fine-grained visual understanding tasks, such as object attribute comprehension. Besides, there have been growing efforts on the evaluations of large vision-language models, but lack of in-depth study of attribute comprehension and the visual language fine-tuning process.
In this paper, we propose to evaluate the attribute comprehension ability of large vision-language models from two perspectives: attribute recognition and attribute hierarchy understanding. We evaluate three vision-language interactions, including visual question answering, image-text matching, and image-text cosine similarity. Furthermore, we explore the factors affecting attribute comprehension during fine-tuning. Through a series of quantitative and qualitative experiments, we introduce three main findings: (1) Large vision-language models possess good attribute recognition ability, but their hierarchical understanding ability is relatively limited. (2) Compared to ITC, ITM exhibits superior capability in capturing finer details, making it more suitable for attribute understanding tasks. (3) The attribute information in the captions used for fine-tuning plays a crucial role in attribute understanding. We hope this work can help guide future progress in fine-grained visual understanding of large vision-language models. The code will be available at {\color{blue} \href{https://github.com/ZHWWWWW/Attribute-Comprehension-of-VLMs}{Attribute-Comprehension-of-VLMs}}.


\end{abstract}
\section{Introduction}
Visual attributes \cite{liang2023hierarchical,pham2021vaw,patterson2016coco,bravo2023ovad,pham2022lsa,zeng2020multi}
are important components of objects, enabling models to describe object information more accurately, which benefits various downstream tasks, including compositional reasoning \cite{gao2023cric} and visual question answering \cite{peng2016vqa,yash2017vqav2}. In recent years, large vision-language models have achieved remarkable achievements, while some works have identified many problems in large vision-language models, such as hallucination issues \cite{li2023evaluating} and social bias \cite{chuang2023debiasing}. A direct question is: How do large-scale vision-language models perform in attribute understanding? Previous works~\cite{yuksekgonul2022aro,yamada2022lemons} find that the model cannot correctly establish the connection between objects and attributes, resulting in degraded visual question answering (VQA) performance. \cite{bravo2023ovad,zhao2022vlcheck} indicate that compared to zero-shot image classification, the absolute performance of attribute recognition is surprisingly low. However, previous studies simply evaluate the large vision-language models on a limited attribute set instead of large-scale attributes in the wild scenario. Besides, understanding attributes not only demands accurate recognition of individual attributes but also comprehending the semantic relationships between attributes, such as hierarchical relationships. 

\begin{figure}[t]
\centering
\includegraphics[width=1\textwidth]{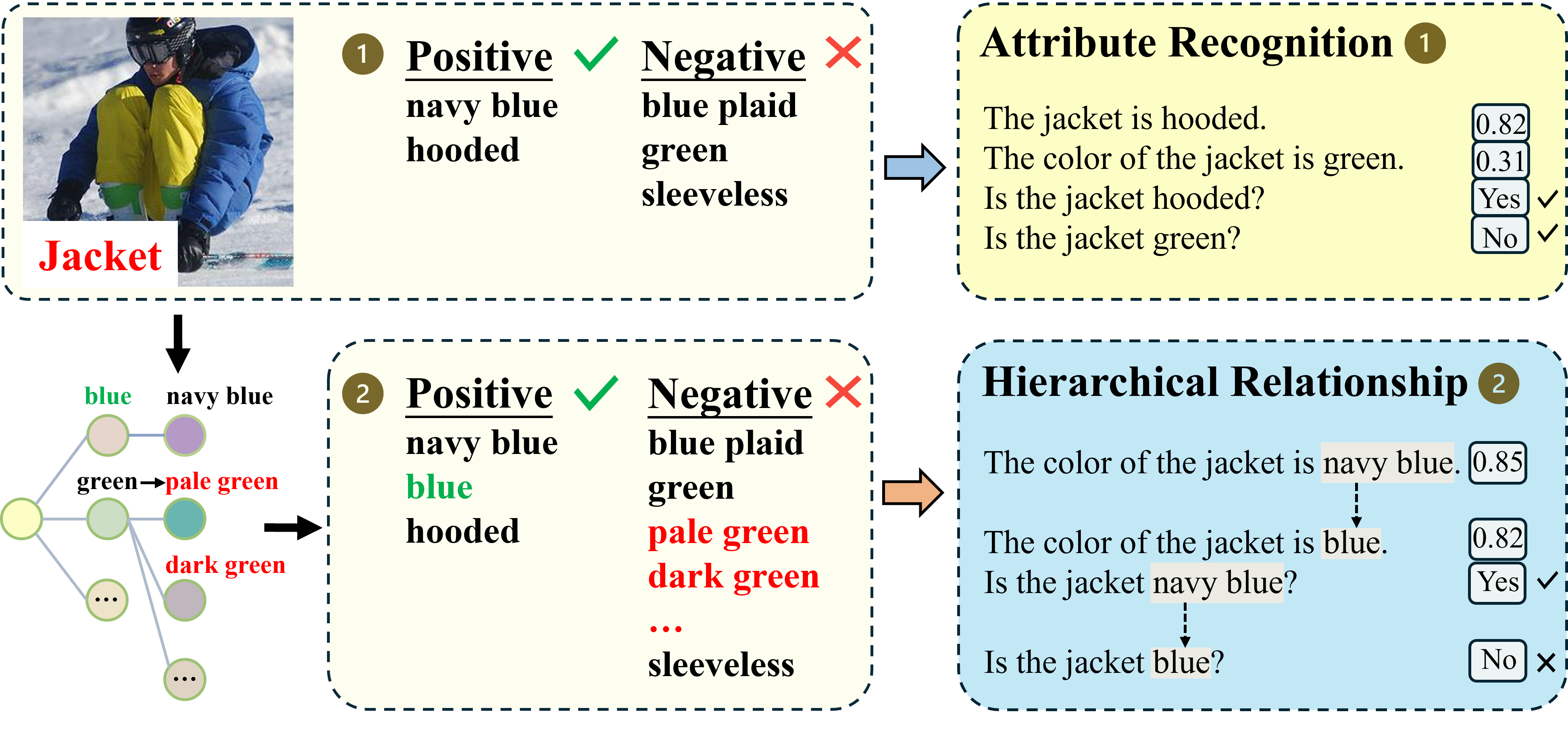}
\caption{Overview of the evaluation process. Original annotations of VAW are used for attribute recognition, while complementary annotations are used to evaluate hierarchical relationship understanding. 
We utilize the attribute tree \cite{liang2023hierarchical} to perform complementation, and the inference results of ITM and ITC are presented as scores. For VQA, the results are presented as either `Yes' or `No'.
}
\label{fig:intro}
\end{figure}

Therefore, in this work, we utilize the VAW dataset \cite{pham2021vaw} to enlarge the scale of wild visual attributes and supplement the hierarchical annotations to evaluate the understanding between attribute semantic relationships. The performance of mainstream large vision-language models in understanding large-scale wild attributes is evaluated from two perspectives: attribute recognition and attribute hierarchical relationship understanding. 
Attribute recognition requires the model to predict the object attributes correctly. While attribute hierarchical relationship understanding requires the model to predict the existence of the parent attribute when it predicts the existence of a child attribute. Similarly, when predicting the absence of a parent attribute, it should also predict the absence of its child attributes. As illustrated in Fig.\ref{fig:intro}, the model predicts the presence of the positive attribute `navy blue', but it outputs `No' for its parent attribute `blue', thus violating the hierarchy.
We conduct evaluation through three approaches: visual question answering (VQA), image-text matching (ITM), and image-text cosine similarity (ITC), reporting attribute understanding ability of different models from two aspects and comparing the distinctions in evaluation methodologies.
Experiments reveal that current large vision-language models have achieved a certain level of attribute comprehension. BLIP \cite{li2022blip}, BLIP2 \cite{li2023blip2}, ALBEF \cite{li2021albef}, and mPLUG \cite{li2022mplug} even surpass supervised models in recognizing tail attributes. 

Secondly, we examine why large vision-language models achieve comparable levels of attribute comprehension capability, even though fine-tuned on coarse image-text pairs rather than high-quality attribute annotations such as VAW \cite{pham2021vaw}. This raises questions about which factors influence the understanding of attributes during fine-tuning. To study the question, we further investigate two aspects: image resolution and the diversity of attribute information in the captions. Our experimental results indicate that the attribute information within the captions plays an important role in the process while image resolution appears to be less important for attribute comprehension.

Overall, our contributions can be summarized as follows: 
(1). We propose an evaluation framework for assessing the attribute understanding capabilities of large vision-language models in terms of attribute recognition and hierarchical relationship understanding.
(2). Through experiments, we find that ITM possesses better detail-capturing ability than ITC and that the two methods respond differently to different templates. ITM requires more detailed templates while ITC performs better when provided only attributes in the templates.
(3). We examine the impact of image resolution as well as attribute information in captions used for fine-tuning and offer insights into the construction of future image-text pairs.

\section{Attribute Understanding Benchmark}

This section presents the evaluation process of attribute recognition and hierarchical relationship understanding. We introduce three evaluation methods: image-text cosine similarity, image-text matching and  visual question answering. All evaluations are conducted on the VAW test set \cite{pham2021vaw}, which encompasses 620 attributes and covers various attribute types, including color, material, shape, size, texture, action, state, and others.

\subsection{Formulation}

Given an image $x$ of an object $o$, $x$ is partially labeled by its corresponding attribute vector $y = \{y_a\}^A_{a=1}$, where $A$ denotes the total number of attributes and $y_a \in \{-1, 0, 1\}$ denotes whether the attribute $a$ is present in the image (`1'), absent (`-1') or unknown (`0'). The positive, negative, and un-annotated attribute sets of $x$ are denoted as $\mathcal{P}_{x} = \{a|y_a = 1\}$, $\mathcal{N}_x = \{a|y_a = -1\}$, and $\mathcal{U}_x = \{a|y_a = 0\}$, respectively. A model with good attribute recognition capability must correctly predict whether an attribute is present in the input image. 

There are also hierarchical relationships between attributes, which can be considered as a directed acyclic graph (DAG) denoted as $\mathcal{T}$ \cite{liang2023hierarchical}. The adjacency matrix of $\mathcal{T}$ is denoted as $\mathcal{R}^{\ast}$, where $\mathcal{R}^{\ast}_{ij}=1$ if and only if attribute $i$ is the parent of attribute $j$. We denote the descendant set of $i$ as $\mathcal{D}_i$, if there is a path from $j$ to $i$ in $\mathcal{T}$, we say $j$ is a descendant of $i$, formulated as $j\in \mathcal{D}_i$. A model understanding attribute hierarchical relationship must predict an attribute's all ancestors present when it predicts the attribute present, and an attribute's all descendants absent when it predicts the attribute absent.

\subsection{Evaluation Aspects}
\label{sec:method}
\subsubsection{Attribute Recognition}
We show the evaluation process in Fig.\ref{fig:intro}. To perform inference, we first populate the template with the attribute and object names to obtain the prompt. Second, the obtained prompt and the image are fed into the vision-language model to get the inference results. This process is repeated three times to perform different evaluation approaches, including ITC, ITM, and VQA. Through ITM and ITC, we can get the inference scores. And it is anticipated that the scores of positive attributes are higher than those of negative attributes. Through VQA, we obtain the answers of ‘Yes’ and ‘No’. And it is expected that the predicted answers for positive attributes are ‘Yes’ and ‘No’ for negative attributes. There are a total of 224,855 positive and negative attributes for 31,819 instances in the test set of VAW. For attribute recognition, we use accuracy to report the metrics for VQA and mean average precision (mAP) to report the metrics for ITM and ITC. 

\subsubsection{Hierarchical Relationship Understanding}
To evaluate the understanding of hierarchical relationships, we first utilize the attribute hierarchy \cite{liang2023hierarchical} to complement the annotations. A complementary example is presented in Fig.\ref{fig:intro}. For the positive attribute `navy blue', we add its parent attribute `blue' to the positive attribute annotation. Since ‘hooded’ has no parent attribute, there is no need to perform a supplement. For the negative attribute `green', we incorporate its descendant attributes into the negative attribute annotation, including `dark green', `pale green', etc.
Similarly, we do not need to operate for `blue plaid' and `sleeveless' since they do not have descendant attributes.
After complementation, we have a total of 396,242 positive and negative attributes. Subsequently, the inference operation is conducted on the complementary dataset. We rely on the constraint violation (CV) and mean average precision post coherence correction (CmAP) metrics to report the models' performance in understanding hierarchical relationships. It is hoped that the scores for parent attributes are higher than those of child attributes. In Fig.\ref{fig:intro}, the score of `navy blue' is higher than `blue', and the model outputs `No' for `blue' while it predicts the presence of `navy blue', which violates hierarchy.

\subsection{Evaluation Methodologies}
\subsubsection{Image-text Cosine Similarity (ITC)}
Many large vision-language models \cite{li2021albef,radford2021clip,li2022blip,li2023blip2,li2022mplug} use image-text cosine similarity as one of the pre-training objectives to align image and text representations in the joint embedding space. Given the image embedding $I$ of $x$ and the text embedding $T$ of attribute prompt $t_a$: `The \{attribute type\} of the \{object\} is \{attribute\}.', we can get the probability of attribute $a$ present in $x$: 
\begin{equation}
p_a = \sigma (cos\left(I, T\right)),
\end{equation}
where $cos(\cdot)$ is the cosine similarity function and $\sigma$ is the sigmoid function to scale the logits into $[0,1]$. This template is shared by ITM and ITC, while we also explore other templates in the ablation study.

\subsubsection{Image-text Matching (ITM)}
It is also a training objective shared by many models, including \cite{li2021albef,li2022blip,li2023blip2,li2022mplug}, which predicts whether a pair of image and text is matched or mismatched. During the training process, hard negative samples from ITC are selected for better fusion between visual and textual information. The ITM head outputs an array with two scores, the first is the mismatch score, and the other is the match score. 
The evaluation process of ITM is similar to ITC, however, embeddings will be entered into the ITM head to get scores:
\begin{equation}
    S = \sigma (F_{itm}(E_{mm}(I, T))),
\end{equation}
where $E_{mm}$ is a multimodal encoder, $F_{itm}$ is the ITM head, and $\sigma$ is the softmax function. Then $p_a$, the probability of attribute $a$ present in $x$ can be obtained by taking the second element of $S$.

\subsubsection{Visual Question Answering (VQA)}
Visual question answering requires the model to answer the question according to the corresponding image. The VQA models used in the experiment are fine-tuned on the VQA dataset \cite{yash2017vqav2}, except for MiniGPT-4 \cite{zhu2023minigpt4}, which only provides the pre-trained model. For fairness, we limit the candidate answers to containing only two answers: ‘Yes’ and ‘No’. The question template is `Is the \{object\} \{attribute\}?', while for material attributes it is `Is the \{object\} made of \{attribute\}?'. 
Since BLIP2 \cite{li2023blip2} does not provide an interface to select from candidate answers, we do not perform evaluations on it.

\subsection{Evaluation Metrics}
To evaluate the attribute understanding capability of different models, we utilize four evaluation metrics.

\textbf{(1) Accuracy}: It serves as an assessment of the overall correctness of a model's responses to questions. 

\textbf{(2) Mean average precision (mAP)}: Mean average precision is a prominent metric particularly applied to attribute prediction and multi-label classification. Due to the partial attributes annotation of VAW, we only consider annotated attributes.

\textbf{(3) Mean average precision post coherence correction (CmAP) \cite{ICLRMBMPatel}}: It post-processes attribute prediction scores to enforce hierarchical constraints within the model’s outputs. Specifically, for a non-leaf attribute, it computes the maximum prediction score among its descendant attributes. Large deviations between mAP and CmAP indicate poorer hierarchical understanding. Given a non-leaf attribute $i$, its post-processing probability:
\begin{equation}
p_i^* = \max\{p_i, \max\{p_j | j \in D_i \}\}.
\end{equation}

\textbf{(4) Constraint violation (CV) \cite{ICLRMBMPatel}} is a metric utilized to assess the hierarchical constraint violations present in a model's predictions. It measures the extent to which model predictions deviate from hierarchical constraints. For ITM and ITC, we compute the CV based on the models' scores
\begin{equation}
CV_{ITM/ITC} = \frac{1}{|\mathcal{K}| \mathcal{|T|}} \sum_{k=1}^{|\mathcal{K}|} \sum_{j \in D_i}  1 \left( p^{(k)}_i - p^{(k)}_j \right) < 0, 
\end{equation}
where $\mathcal{|K|}$ is the number of test images and $\mathcal{|T|}$ is the number of edges in the attribute tree $\mathcal{T}$. For VQA, we calculate the CV based on the models' answers. We use $T^*$ to represent the number of edges present in the image after complementation in the manner described in \ref{sec:method}. While $F^*$ is to represent the number of edges that violate hierarchy, then we have:
\begin{equation}
CV_{VQA} = \frac{1}{|\mathcal{K}|}\sum_{k=1}^{|\mathcal{K}|}\frac{F^*}{T^*}.
\end{equation}

\subsection{Comparison Methods}

The models compared in our experiments are as follows. We list the corresponding model weights of evaluated large vision-language models in Table \ref{tab:weight parameters}.

\textbf{ResNet-Baseline} \cite{patterson2016coco} provides a baseline for object attribute recognition.
\textbf{VAW} \cite{pham2021vaw} proposes a strong baseline model along with supervised contrastive learning using negative-label expansion.
\textbf{CLIP} \cite{radford2021clip} is trained on 400 million image-text pairs, using a text encoder and a visual encoder to map images and texts to the same feature space.
\textbf{ALBEF} \cite{li2021albef} employs contrastive alignment of image and text representations before fusion, enabling robust vision-language learning without needing bounding box annotations or high-resolution images. 
\textbf{BLIP} \cite{li2022blip} adeptly handles understanding and generation tasks, leveraging noisy web data through caption bootstrapping. 
\textbf{BLIP2} \cite{li2023blip2} bridges frozen pre-trained image encoders and language models by training a lightweight querying transformer.
\textbf{mPLUG} \cite{li2022mplug} proposes a cross-modal skipped-connected network to address the problem of information asymmetry between image and text and improve computational efficiency.
\textbf{MiniGPT-4} \cite{zhu2023minigpt4} aligns a frozen text encoder and a frozen LLM using one projection layer to achieve advanced multi-modal capabilities.

\begin{table}[t]
\begin{center}
\caption{This table represents the weight parameters and corresponding datasets for our evaluated models. 
A: Conceptual Captions \cite{sharma2018conceptual}; B: SBU Captions \cite{ordonez2011im2text}; C: COCO \cite{lin2014microsoft}; D: Visual Genome \cite{krishna2017visual}; E: Conceptual 12M \cite{changpinyo2021conceptual}; F: VQA \cite{antol2015vqa}; G: LAION400M \cite{schuhmann2021laion}; H: Flickr30K \cite{plummer2015flickr30k}.} \label{tab:weight parameters}
\resizebox{1\textwidth}{!}{
\begin{tabular}{|l|c|c|c|ccc|}
  \hline
  \multicolumn{1}{|c|}{\multirow{2}{*}{Models}}  &Pre-training  &Weight  &Fine-tuning  & VQA & ITM & ITC \\ &datasets & & datasets & & & \\ \hline
  
 CLIP \cite{radford2021clip} &WebImageText   & ViT-B/32(151.3M) &$-$  &  &  &\checkmark  \\ \hline
 
 \multirow{2}{*}{ALBEF\cite{li2021albef}} &A B C &albef\_vqav2\_lavis(580.7M)   &F  & \checkmark &  &\\ 
 
 &D E  &albef\_coco\_retrieval\_lavis(419.1M)  &C  &   & \checkmark  & \checkmark \\ \hline
 
 \multirow{2}{*}{BLIP \cite{li2022blip}} &B C D&model\_base\_vqa\_capfilt\_large(361.5M)
 & F   &\checkmark  & & \\
 
  &E G&model\_base\_retrieval\_coco(223.7M)
 &C &  &\checkmark  &\checkmark \\ \hline

  BLIP2 \cite{li2023blip2} & A B C D E G & blip2\_finetune\_coco(1173.2M)  & C &  &\checkmark &\checkmark  \\ \hline

  \multirow{2}{*}{mPLUG \cite{li2022mplug}} &A B C & mplug\_visual-question-answering\_coco\_large\_en(1494.3M) &F &\checkmark &  & \\
 
  &D E &mplug\_image-text-retrieval\_flickr30k\_large\_en(1217.8M)  &H &   &\checkmark &\checkmark \\ \hline

  MiniGPT-4 \cite{zhu2023minigpt4} & A B E G & pretrained\_minigpt4\_7b(7832.6M)  &$-$  & \checkmark  &   &   \\ 
 \hline
\end{tabular}
}
\end{center}
\end{table}

\section{Experimental Results}
In this section, we show the experimental results of our proposed benchmark and compare different models through a series of analyses.
\begin{table}[t]
\caption{Comparison of attribute recognition ability with the close-set models on the VAW dataset \cite{pham2021vaw}. The best overall results are in bold. The best results in close-set models, ITC, and ITM are highlighted in \colorbox{T1_green!40}{Green}, \colorbox{red!20}{Red}, and \colorbox{T1_blue}{Blue}, respectively.}
\label{tab1}
\small
\resizebox{1\textwidth}{!}{
\begin{tabular}{|l|c|ccc|cccccccc|}
\hline
\multicolumn{1}{|c|}{\multirow{2}{*}{Methods}} & \multirow{2}{*}{\begin{tabular}[c]{@{}c@{}}Overall \\ (mAP)\end{tabular}} & \multicolumn{3}{c|}{Class imbalance(mAP)}        & \multicolumn{8}{c|}{Attribute types (mAP)}                                                                                             \\
\multicolumn{1}{|c|}{}                         &                                                                           & Head           & Medium         & Tail           & Color          & Material       & Shape          & Size           & Texture        & Action         & State          & Others         \\ \hline

\multicolumn{13}{|l|}{\textit{\color{blue}{Close-set models}}}   \\ \hline

ResNet-Baseline \cite{patterson2016coco} & 63.0 & 71.1 & 59.4 & 43.0 & 58.5 & 66.3 & 65.0 & 64.5 & 63.1 & 53.1 & $-$ & 66.7 \\

VAW \cite{pham2021vaw} & {\cellcolor{T1_green!40}{\textbf{68.3}}} & {\cellcolor{T1_green!40}{\textbf{76.5}}} & {\cellcolor{T1_green!40}{64.8}} & {\cellcolor{T1_green!40}{48.0}} & {\cellcolor{T1_green!40}{\textbf{70.4}}} & {\cellcolor{T1_green!40}{\textbf{75.6}}} & {\cellcolor{T1_green!40}{\textbf{68.3}}} & {\cellcolor{T1_green!40}{\textbf{69.4}}} & {\cellcolor{T1_green!40}{\textbf{68.4}}} & {\cellcolor{T1_green!40}{60.7}} & $-$ & {\cellcolor{T1_green!40}{\textbf{69.5}}} \\

\hline

\multicolumn{13}{|l|}{\textit{\color{blue}Open-set models image-text cosine similarity}}   \\ \hline

CLIP \cite{radford2021clip}   &  42.1 &     43.6 &    43.2 &  33.2&     33.0 &33.6 &   36.8&  42.3&     43.3&     38.2&     41.7 & 47.1   \\

ALBEF \cite{li2021albef}   &49.0       & 48.5         &51.8    & 42.5 &38.2  & 37.9 & 42.8&          43.5&    44.7 & 54.6        &46.6         &54.2  \\

BLIP \cite{li2022blip}       &52.2      & 52.5         &54.4          &44.5           & 44.2          &44.6  &44.9&45.5    &48.6       &53.2           &49.2                      &57.2           \\

BLIP2 \cite{li2023blip2}    &{\cellcolor{red!20}{58.0}}        &{\cellcolor{red!20}{56.4}}           & {\cellcolor{red!20}{60.9}}          &{\cellcolor{red!20}{55.3}}           &{\cellcolor{red!20}{49.9}}           &{\cellcolor{red!20}{52.9}}           &{\cellcolor{red!20}{53.1}}           & {\cellcolor{red!20}{47.0}}          &{\cellcolor{red!20}{55.4}}         & {\cellcolor{red!20}{\textbf{63.9}}}         &{\cellcolor{red!20}{53.1}}          & {\cellcolor{red!20}{62.0}}        \\

mPLUG \cite{li2022mplug} &45.3                                                                      & 45.0          & 47.4          & 40.4          & 35.5          &          34.0&           41.0&  41.4&           39.3&         45.5&           46.01& 50.7 \\ \hline

\multicolumn{13}{|l|}{\textit{\color{blue}{Open-set models image-text matching}}}   \\ \hline

ALBEF \cite{li2021albef}     & 56.3      & 55.9       & 58.9   & 49.7 & 51.6        &47.5     & 47.6         &45.0      &  54.8    & 59.3    & 51.4                 &60.7  \\

BLIP \cite{li2022blip}       &      59.1&          57.9&          62.1&           54.3&           {\cellcolor{T1_blue}{57.5}}&  51.7&         49.0 &           {\cellcolor{T1_blue}{47.2}} &           58.1 &           57.1&           52.5&63.7           \\

BLIP2 \cite{li2023blip2}    &{\cellcolor{T1_blue}{62.7}}        &{\cellcolor{T1_blue}{60.5}}           &{\cellcolor{T1_blue}{\textbf{65.7}}}           &{\cellcolor{T1_blue}{\textbf{61.5}}}          & 57.2          &{\cellcolor{T1_blue}{56.1}}           &{\cellcolor{T1_blue}{57.5}}           &46.2           &{\cellcolor{T1_blue}{59.4}}         & {\cellcolor{T1_blue}{63.2}}          &55.2        &{\cellcolor{T1_blue}{67.9}}         \\

mPLUG \cite{li2022mplug} & 60.0                                                                     & 58.8          & 62.9          & 55.6          &56.0           & 53.6         & 49.7          &45.8  & 54.7         &61.2        & {\cellcolor{T1_blue}{\textbf{58.5}}}           &64.3  \\

\hline
\end{tabular}
}
\end{table}

\begin{figure}[ht]
  \begin{minipage}[c]{0.55\textwidth}
     \centering
        \includegraphics[width=\linewidth]{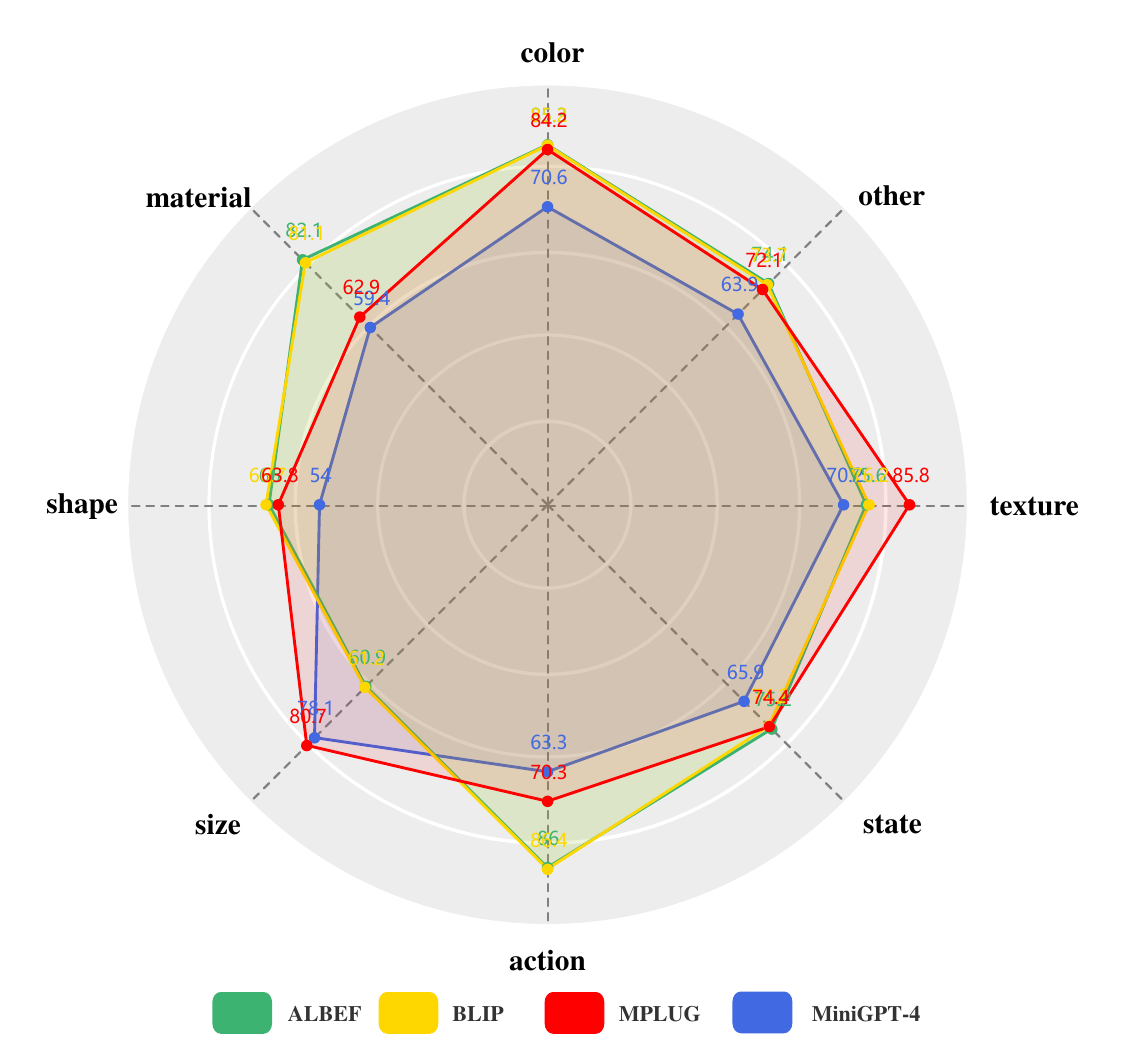}
        \caption{Attribute recognition accuracy of different models. The performance is assessed across eight attribute types within the VAW dataset.}
        \label{fig:vqa}
  \end{minipage}
  \begin{minipage}[c]{0.45\textwidth}
     \centering
     \captionof{table}{Experimental results of hierarchical understanding ability of different models. The computation of CV for the close-set models is consistent with that for ITM and ITC, so we do not compare it with VQA. The best results are in bold.}
    \label{tab2}
    \resizebox{\linewidth}{!}{
    \begin{tabular}{|l|c|cc|cc|}
      \hline
    
      Method & \multicolumn{3}{c|}{CV $\downarrow$} & \multicolumn{2}{c|}{CmAP $\uparrow$} \\ \hline

      \multicolumn{6}{|l|}{\textit{\color{blue}{Close-set models}}}   \\ \hline
    
      VAW \cite{pham2021vaw}  & $-$ & \multicolumn{2}{c|}{\textbf{23.7}}  & \multicolumn{2}{c|}{\textbf{68.3}}\\ \hline
    
      \multicolumn{1}{|l|}{\textit{\color{blue}{Open-set models}}} &  VQA & \multicolumn{1}{c}{ITM} & \multicolumn{1}{c|}{ITC} &
      \multicolumn{1}{c}{ITM} & \multicolumn{1}{c|}{ITC} \\ \hline
      
      CLIP \cite{radford2021clip} & $-$ & $-$ & 58.2 &$-$  & 47.9 \\
      ALBEF \cite{li2021albef} & \textbf{8.24}  & 43.1  & 45.0  &60.7  & 54.3 \\
      BLIP \cite{li2022blip} & 8.75  &34.7  & 34.1  & 63.3  & 57.2 \\
      BLIP2 \cite{li2023blip2} & $-$  & 41.5  & 42.7  & 66.7  &62.5 \\
      mPLUG \cite{li2022mplug} & 9.84 & 48.15  & 57.4  & 64.1  & 50.9\\
      MiniGPT-4 \cite{zhu2023minigpt4} & 11.37 & $-$ & $-$  & $-$ & $-$ \\
      \hline
    \end{tabular}
    }
  \end{minipage}

\end{figure}

\subsection{Comparison with Close-Set Models}
We demonstrate the attribute understanding ability of different models in terms of attribute recognition and hierarchical relationship understanding, each of which is evaluated in three ways: VQA, ITM, and ITC. For attribute recognition results, Table \ref{tab1} presents a comparison between close-set models and the open-set models using ITC and ITM. We find that large vision-language models exhibit substantial attribute comprehension capabilities. For instance, the overall mAP of BLIP2's ITM is very close to the ResNet-Baseline, which is trained on the VAW dataset. Additionally, for tail attributes, the performance of mPLUG, BLIP, and BLIP2 using ITM is even much better than supervised models, indicating that the data imbalance of the training set impairs the performance of supervised models. In contrast, large vision-language models with strong generalization capabilities can mitigate this issue. 

Fig.\ref{fig:vqa} represents the attribute recognition results of VQA. Among them, ALBEF, BLIP, and mPLUG obtain comparable recognition results, but mPLUG and the previous two models have differences in recognizing different attribute types. Table \ref{tab2} shows the results of hierarchical relationship understanding capability, we find large vision-language models can comprehend certain hierarchical relationships. However, there are some deviations between CmAP and mAP. Besides, the CV metrics for ITM and ITC also have a significant gap compared to the supervised model, which demonstrates less favorable hierarchical understanding capability since the scores of some parent attributes are lower than those of their descendent attributes. The CV metrics for VQA seem to be better since it only predicts the presence and absence of an attribute without the limitation of scores. In both attribute recognition and hierarchical relationship understanding evaluations, BLIP2 achieves the best ITM and ITC inference results.

\begin{figure}[t]
\centering
\includegraphics[width=0.9\textwidth]{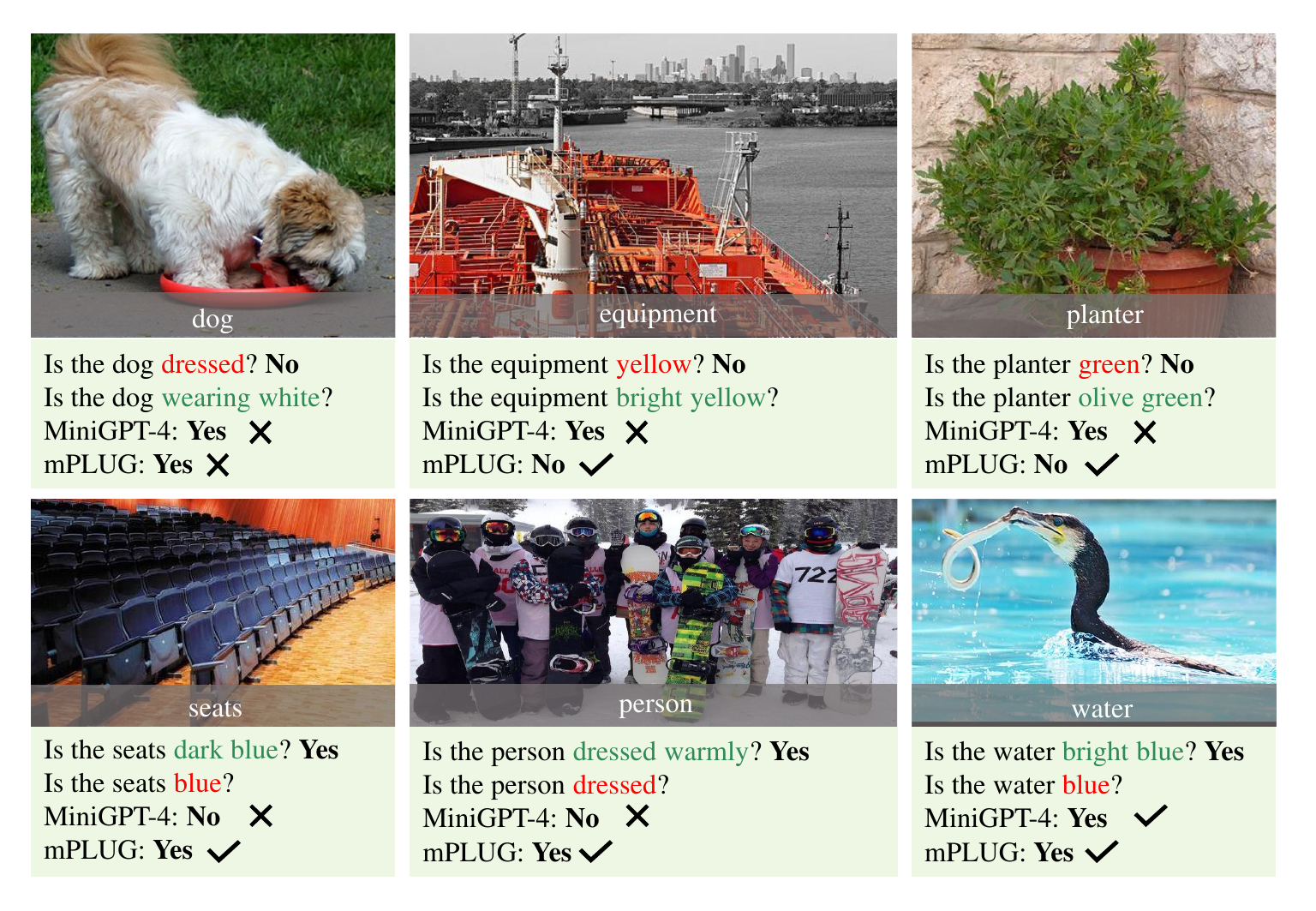}
\caption{Hierarchical relationship understanding comparison between mPLUG and MiniGPT-4, where the parent attributes are highlighted in red and the children attributes are highlighted in green.}
\label{fig:hierarchy}
\end{figure}

\subsection{Analysis}

\subsubsection{What Affects Hierarchical Relationship Understanding.} According to Table \ref{tab2}, mPLUG understands hierarchical relationships better than MiniGPT-4. Through the visualization results in Fig.\ref{fig:hierarchy}, we identify two factors that may influence hierarchical relationship understanding. First, other objects in the image contain the attribute. Like the `olive green planter' in Fig.\ref{fig:hierarchy}, though both models can correctly determine that green is a negative attribute of the planter, MiniGPT-4 mistakenly identifies it as olive green, which violates hierarchy. The question is about the planter rather than the plant, but it fails to locate the right area when recognizing `olive green'. Second, the attribute phrase is not present on the object, but the attribute contained in the phrase is present on the object. Such as the `wearing white dog' in Fig.\ref{fig:hierarchy}, both models misidentify when recognizing whether the dog is wearing white or not, while they can distinguish that the dog is not dressed. It is probably because the dog is white and they cannot understand `wearing white'. These visualizations suggest that sometimes models behave like `bags-of-words' \cite{yuksekgonul2022aro} and lack accurate semantic comprehension, thus impairing attribute understanding.

\begin{figure*}[t]
\centering
\includegraphics[width=1\textwidth]{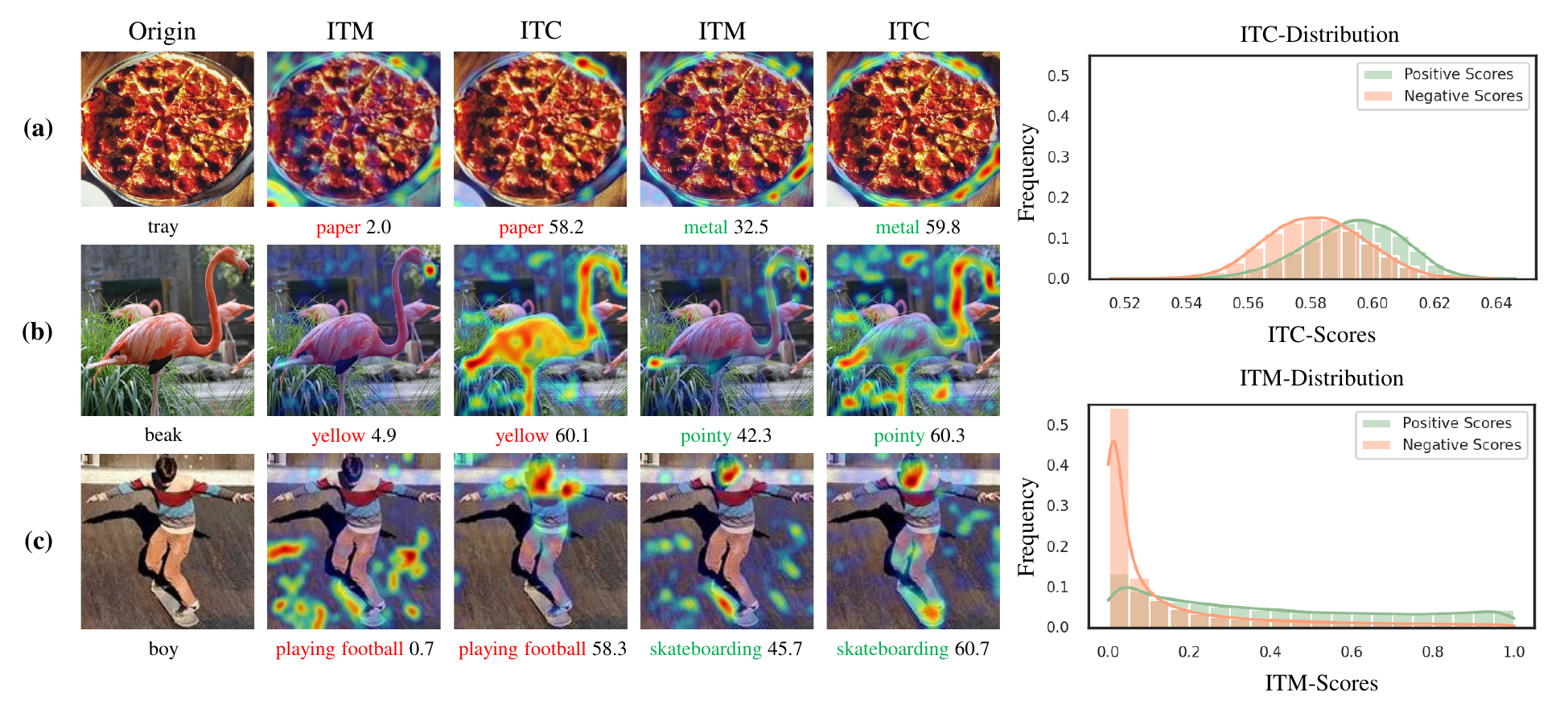}
 \caption{\textbf{Left}: Comparison between Grad-CAM\cite{selvaraju2017grad} visualizations for BLIP ITM and ITC, corresponding to the prompt. Scores are displayed below the picture. Prompts: (a). Negative: The material of the tray is \textcolor{red}{paper}. Positive: The material of the tray is \textcolor{pos_green}{metal}. (b). Negative: The color of the beak is \textcolor{red}{yellow}. Positive: The shape of the beak is \textcolor{pos_green}{pointy}. (c). Negative: The sport activity of the boy is \textcolor{red}{playing football}. Positive: The sport activity of the boy is \textcolor{pos_green}{skateboarding}. Scores are present below the picture. \textbf{Right}: The distribution of positive and negative attribute prediction scores for BLIP \cite{li2022blip}.}
 \label{fig:gradcam_scores}
\end{figure*}

\subsubsection{Effects of Different Evaluation Methods.} In Table \ref{tab1}, we observe that the mAP values of ITM are significantly higher than those of ITC for the same model. This phenomenon can be explained by analyzing the score distribution. Here, we refer to \cite{huang2023open} to perform analysis. Fig.\ref{fig:gradcam_scores} Right shows that ITM has a distinct threshold between positive and negative attributes, with most negative attribute scores concentrated in the low-score region. In contrast, ITC lacks a clear boundary between positive and negative attributes, leading to a lower mAP. 
To further analyze the results, we use Grad-CAM \cite{selvaraju2017grad} to visualize the interpretable components that contribute to the prediction and the degree of contribution of each feature. 
The visualizations in Fig.\ref{fig:gradcam_scores} Left show that both ITM and ITC can correctly locate the target object when provided with a prompt of a positive attribute. However, when recognizing negative attributes that are not present on the object, ITC only concentrates on the entire object while ITM could explore relevant regions of the object that may be associated with that attribute. 

As shown in Fig.\ref{fig:gradcam_scores} left (c), ITC still focuses on the boy when provided with the negative attribute `playing football', whereas ITM attempts to explore the specified attribute and successfully directs attention to the boy's feet. Additionally, we find that the scores for positive and negative attributes vary widely for ITM, but the difference is not obvious for ITC. For example, in Fig.\ref{fig:gradcam_scores} left (c), the difference between positive and negative scores for ITM is 45, while it becomes 2.4 for ITC. The score disparity suggests that ITM is more sensitive to attribute information. 
In contrast, ITC is limited to object-level recognition and lacks the ability for finer-grained attribute analysis. This may be due to the hard negative sampling strategy in BLIP's ITM training \cite{li2021albef,li2022blip,li2022mplug}, which enables ITM to capture more nuanced information and achieve better performance in attribute-level analysis.

\begin{table}[ht]
        \caption{The mAP results for different prompt templates on the VAW dataset. \textbf{T1}: `The object is \{attribute\}.'; \textbf{T2}: `The \{object\} is \{attribute\}.'. We highlight the best template for ITC in {\cellcolor{T1_blue}{blue}} and for ITM in {\cellcolor{red!18}{red}}.}
        \centering
        \label{tab6}
        \small
        \resizebox{0.7\textwidth}{!}{
        \begin{tabular}{|c|c|c|cc|cc|cc|}
        \hline
        \multirow{2}{*}{Template} & \multirow{2}{*}{Object} & CLIP \cite{radford2021clip} & \multicolumn{2}{c|}{BLIP \cite{li2022blip}} & \multicolumn{2}{c|}{mPLUG \cite{li2022mplug}}  & \multicolumn{2}{c|}{ALBEF \cite{li2021albef}}\\ \cline{3-9}
        
            &        & ITC          & ITC           & ITM           & ITC            & ITM          & ITC & ITM \\ \hline
        T1 &   & {\cellcolor{T1_blue}{45.7}}        & 53.8         & 54.1         & {\cellcolor{T1_blue}{46.9}}          & 57.9   & {\cellcolor{T1_blue}{51.4}}  & 49.3      \\ 
        T2 & \checkmark  & 43.2        & {\cellcolor{T1_blue}{53.9}}         & {\cellcolor{red!18}{61.2}}         & 46.2          & {\cellcolor{red!18}{61.1}}   & 50.1   & {\cellcolor{red!18}{58.2}}      \\ \hline
        \end{tabular}
        }
\end{table}

\section{Ablation Study}
The evaluation results indicate that the fine-tuned vision-language models possess certain attribute comprehension capabilities. To further explore the factors affecting the fine-tuning performance and the impact of different prompt templates, we conduct ablation experiments.

\subsection{Effects of Different Templates}
In previous ITC and ITM experiments, we use the template that includes attribute types, object names, and attribute names. To investigate the impact of different prompting templates, we attempt two additional templates. The first template includes only attributes: `The object is \{attribute\}'; the second template includes both object and attribute names: `The \{object\} is \{attribute\}'. As shown in Table \ref{tab6}, ITC and ITM respond differently to various templates. For ITC, providing only attribute information often yields the best attribute recognition performance. However, for ITM, solely providing attributes is not sufficient; it requires the addition of attribute-dependent information, i.e., the object, to achieve better attribute recognition. This indicates that ITM possesses better detail-capturing capabilities, and thus, it performs better when more detailed descriptions are provided. In contrast, ITC lacks compositional understanding, so adding additional object information can introduce interference, leading to reduced performance in recognizing attributes. However, ITM requires more time and computational resources for inference compared to ITC. Therefore, enhancing the effectiveness of contrastive learning remains crucial. 

\subsection{Exploratory Experiments on Factors Affecting the Performance of Fine-tuning}
\begin{table}[ht]
\caption{Evaluation results of different image resolution fine-tuning based on ALBEF \cite{li2021albef}. For text retrieval (TR) and image retrieval (IR), we report the average of R@1, R@5, and R@10. The best results are in bold and we also highlight the results inferenced with the officially provided model in red.}\label{tab3}
\centering
\resizebox{0.7\textwidth}{!}{
\begin{tabular}{|c|cc|cc|}
\hline
\multicolumn{1}{|c|}{\multirow{2}{*}{Image Res}} & \multicolumn{2}{c|}{COCO Retrieval(Avg(R@X))}     & \multicolumn{2}{c|}{Attribute Recognition(mAP)} \\ 
 &TR & IR & ITC & ITM \\
\hline
256 $\times$ 256 & 88.6 & 77.2 & 48.5 & 56.5\\
384 $\times$ 384 & 89.6 (\textcolor{red}{89.6}) & 78.6 (\textcolor{red}{78.5}) & \textbf{49.6} (\textcolor{red}{49.0}) & \textbf{56.6} (\textcolor{red}{56.3})\\
480 $\times$ 480 & \textbf{89.9} & \textbf{79.0} & 48.9 & 55.9\\
\hline
\end{tabular}
}
\end{table}
\subsubsection{Image resolution.} We find that during fine-tuning, most models tend to increase image resolution. For instance, ALBEF \cite{li2021albef} takes random image crops of resolution 256 $\times$ 256 as input during pretraining, while during fine-tuning, it gets boosted to 384 $\times$ 384. Similarly, BLIP \cite{li2022blip} grows from 224 $\times$ 224 to 384 $\times$ 384, and for the VQA task, the resolution is 480 $\times$ 480.
Considering that higher resolution allows models to capture more subtle visual details, which may be beneficial for attribute understanding, we use ALBEF \cite{li2021albef} to explore the effect of image resolution on attribute comprehension during fine-tuning.

The results are shown in Table \ref{tab3}. We use the model obtained by fine-tuning on COCO \cite{lin2014microsoft} dataset for the retrieval task. Therefore, we report the text retrieval (TR) and image retrieval (IR) results on the COCO dataset, as well as the attribute recognition results on the VAW test set. To maintain consistency, we reproduce it with the same image resolution as in the original paper (384 $\times$ 384). Then, we decrease or increase the resolution while keeping other parameters constant during the process. From Table \ref{tab3}, we can see that as the image resolution increases, the performance on the retrieval task becomes better, while the attribute recognition ability initially rises and then falls. These results demonstrate that image resolution is not the main factor contributing to attribute comprehension capabilities since no significant fluctuations are detected when it is increased from 256 $\times$ 256 to 384 $\times$ 384, but it is effective for retrieval tasks.

\subsubsection{Attribute Density in Captions.}  Previous work on attribute predictions requires datasets with high-quality attribute annotations, such as VAW \cite{pham2021vaw}. However, the dataset used for finetuning is composed of image-text pairs, which do not provide a detailed description of the attributes in the image. To investigate the effect of captions in the training set on attribute comprehension, we count the overlapping attributes of the COCO training set with the VAW dataset. The VAW dataset has a total of 620 attributes, 605 of which are included in the training set of COCO. After removing the 15 non-overlapping attributes, the mAPs of ALBEF are increased, while for ITC it is 49.0 $\rightarrow$ 49.2 and for ITM it is 56.3 $\rightarrow$ 56.5. 
This preliminary outcome indicates that the diversity of attributes in the training set may influence the ability to understand attributes. To further validate this hypothesis, we conduct fine-tuning experiments on ALBEF by removing attributes of certain types from captions. 
Two attempts are made, including `Color Absent' and `Material Absent', which correspond to deleting color information and material information, respectively.

We utilize `bert-base-uncased' for the deletion operation. 
For example, to delete color information from COCO, first, the color attributes of VAW and the captions of the COCO training set are encoded to obtain the input ids for attributes and captions. Secondly, those in the input ids for captions that overlap with the input ids for attributes are removed. Thirdly, the input ids for captions are decoded and the first letter of each resulting caption is capitalized to obtain a new caption. After the process, the original caption `Two people are riding a red bike down the street.' is transformed into `Two people are riding a bike down the street.', where `red' is removed. However, it is not possible to remove all the color and material information from the captions, such as `reddish', which is not present in the VAW dataset. Furthermore, the plural form of `fabric', `fabrics', cannot be removed since it has a different input id. Therefore, there is still color and material information after deletion, but the diversity is decreased.

\begin{table}[t]
\caption{Comparison between `color absent', `material absent', and our reproduced results on attribute recognition. All experiments are conducted on ALBEF \cite{li2021albef}. The changes of color and material types are highlighted in blue and green, respectively.}
\label{tab4}
\small
\resizebox{1\textwidth}{!}{
\begin{tabular}{|l|c|ccc|cccccccc|}
\hline
\multicolumn{1}{|c|}{\multirow{2}{*}{Methods}} & \multirow{2}{*}{\begin{tabular}[c]{@{}c@{}}Overall \\ (mAP)\end{tabular}} & \multicolumn{3}{c|}{Class imbalance(mAP)}        & \multicolumn{8}{c|}{Attribute types (mAP)}                                                                                             \\
\multicolumn{1}{|c|}{}                         &                                                                           & Head           & Medium         & Tail           & Color          & Material       & Shape          & Size           & Texture        & Action         & State          & Others         \\ \hline
\multicolumn{13}{|l|}{\textit{\color{blue}Open-set models image-text cosine similarity}}   \\ \hline

Color Absent   &46.9      & 47.0   &49.6  &38.2  & {\cellcolor{T1_blue}{32.3}}  & 36.5 & 40.9&          42.7&    44.4 & 51.9         &45.4         &52.8  \\
Material Absent   &48.7      & 48.3   &51.4  &41.4  & 38.2  & {\cellcolor{T1_green!40}{36.0}} & 42.0&          43.4&    45.9 & 52.9         &47.0         &54.0  \\
Reproduced   &49.6       & 49.2         &52.3    & 42.7 & {\cellcolor{T1_blue}{39.0}}  & {\cellcolor{T1_green!40}{38.6}} & 42.5 & 43.6    &45.5  & 54.0       &46.8        &55.1  \\
\hline

\multicolumn{13}{|l|}{\textit{\color{blue}{Open-set models image-text matching}}}   \\ \hline

Color Absent     & 55.0      & 54.5       & 57.7   & 48.6 & {\cellcolor{T1_blue}{47.6}}        &45.7     & 47.1         &44.3          & 53.2    & 59.4          &  49.7       &59.9  \\
Material Absent     & 56.3      & 55.8       & 59.1   & 49.2 & 51.6        &{\cellcolor{T1_green!40}{46.5}}     & 46.9         &44.8          & 55.7    & 59.3          &  51.7       &60.8  \\
Reproduced     & 56.6      & 56.1      & 59.4   & 50.0 & {\cellcolor{T1_blue}{51.5}}        &{\cellcolor{T1_green!40}{48.2}}     & 47.2         &45.1          & 55.5    & 59.2          &  51.6      &61.2  \\
\hline

\end{tabular}
}
\end{table}


We compare the results of removing color and material attributes with our reproduced results in Table \ref{tab4}. Significant declines are observed for the AP of color and material type. This evidence supports the hypothesis that during the fine-tuning process, attribute information contained in the captions enhances the model's attribute understanding capabilities, and the diversity of this attribute information plays an important role in the process.
It is also notable that the removal of color information results in a notable reduction in other attribute types, such as `material'.
This could be attributed to the material-related modifications that occur during the process of removing color information, which in turn leads to a decrease in metrics. 

\section{Conclusion}
This study examines the ability of large vision-language models to understand object attributes from two aspects: attribute recognition and attribute hierarchical relationship understanding. Our experiments highlight the differences between ITM and ITC in terms of attribute understanding and the impact of different prompt templates. Furthermore, we investigate the factors that affect attribute comprehension during the fine-tuning process, excluding factors that have a minor impact, such as image resolution. We also confirm the significance of diversity of attribute information in captions. 
With the advancement of large vision-language models, it has become easier to generate image-text pair datasets. 
This may make it possible to specify dimensions to be enriched to obtain more appropriate datasets for customizing models.


%
%
%
%

\bibliographystyle{splncs04}
\bibliography{prcv2024}






  
\end{document}